\documentclass{article}

\usepackage{microtype}
\usepackage{graphicx}
\usepackage{subfigure}
\usepackage{booktabs} 
\usepackage[table]{xcolor}
\usepackage[T1]{fontenc}

\usepackage{hyperref}

\usepackage[accepted]{icml2025}

\usepackage{amsmath}
\usepackage{amssymb}
\usepackage{mathtools}
\usepackage{amsthm}

\usepackage[capitalize,noabbrev]{cleveref}

\theoremstyle{plain}

\theoremstyle{definition}

\theoremstyle{remark}

\usepackage[textsize=tiny]{todonotes}

\icmltitlerunning{Deploying Geospatial Foundation Models in the Real World: Lessons from WorldCereal}

\begin{document}

\twocolumn[
\icmltitle{Deploying Geospatial Foundation Models in the Real World: Lessons from WorldCereal}



\icmlsetsymbol{equal}{*}

\begin{icmlauthorlist}
\icmlauthor{Christina Butsko}{equal,vito}
\icmlauthor{Kristof Van Tricht}{equal,vito}
\icmlauthor{Gabriel Tseng}{equal,mila,mcgill}
\icmlauthor{Giorgia Milli}{vito}
\icmlauthor{David Rolnick}{mila,mcgill}
\icmlauthor{Ruben Cartuyvels}{kul,esa}
\icmlauthor{Inbal Becker Reshef}{unistr}
\icmlauthor{Zoltan Szantoi}{esa}
\icmlauthor{Hannah Kerner}{ari}
\end{icmlauthorlist}

\icmlaffiliation{vito}{VITO, Belgium}
\icmlaffiliation{mila}{Mila -- Quebec AI Institute, Canada}
\icmlaffiliation{mcgill}{McGill University, Canada}
\icmlaffiliation{kul}{KU Leuven, Belgium}
\icmlaffiliation{unistr}{University of Strasbourg, France}
\icmlaffiliation{esa}{European Space Agency, France}
\icmlaffiliation{ari}{Arizona State University, United States of America}

\icmlcorrespondingauthor{Christina Butsko}{christina.butsko@gmail.com}
\icmlcorrespondingauthor{Kristof Van Tricht}{kristof.vantricht@vito.be}
\icmlcorrespondingauthor{Gabriel Tseng}{gabrieltseng95@gmail.com}

\icmlkeywords{Machine Learning, ICML}

\vskip 0.3in
]



\printAffiliationsAndNotice{\icmlEqualContribution} 

\begin{abstract}
The increasing availability of geospatial foundation models has the potential to transform remote sensing applications such as land cover classification, environmental monitoring, and change detection. Despite promising benchmark results, the deployment of these models in operational settings is challenging and rare. Standardized evaluation tasks often fail to capture real-world complexities relevant for end-user adoption such as data heterogeneity, resource constraints, and application-specific requirements. This paper presents a structured approach to integrate geospatial foundation models into operational mapping systems. Our protocol has three key steps: defining application requirements, adapting the model to domain-specific data and conducting rigorous empirical testing. Using the Presto model in a case study for crop mapping, we demonstrate that fine-tuning a pre-trained model significantly improves performance over conventional supervised methods. Our results highlight the model's strong spatial and temporal generalization capabilities. Our protocol provides a replicable blueprint for practitioners and lays the groundwork for future research to operationalize foundation models in diverse remote sensing applications. Application of the protocol to the WorldCereal global crop-mapping system showcases the framework's scalability.
\end{abstract}

\section{Introduction}

A growing number of geospatial and remote sensing foundation models have emerged in recent years, such as Presto \cite{tseng2024lightweightpretrainedtransformersremote}, ScaleMAE \cite{Reed_Gupta_Li_Brockman_Funk_Clipp_Keutzer_Candido_Uyttendaele_Darrell_2023}, SatMAE \cite{cong2022satmae}, AnySat \cite{astruc2024anysatearthobservationmodel}, CROMA \cite{fuller2023croma}, SkySense \cite{guo2024skysense}, and others \cite{bastani2023satlaspretrainlargescaledatasetremote, smith2024earthpttimeseriesfoundation, irvin2023usatunifiedselfsupervisedencoder, xiong2024allunifiedfoundationmodels, wang2024mtpadvancingremotesensing, jiang2024lemevitefficientvisiontransformer, 10697182, szwarcman2025prithvieo20versatilemultitemporalfoundation, li2024seamomultiseasonalmultimodalremote}. Each reports impressive performance and promises to revolutionize a wide range of applications in remote sensing, such as land cover classification, change detection and environmental monitoring. Leveraging large-scale self-supervised pre-training, these models promise to enable generalization over diverse global patterns and data distributions. The encouraging results on diverse benchmarks (e.g., GeoBench \cite{lacoste2023geo}, Pangaea \cite{marsocci2024pangaea}, PhilEO Bench \cite{fibaek_phileo}) give us hope that foundation models can play a pivotal role in addressing real-world Earth observation challenges.

Despite promising results on benchmarks, the integration of geospatial foundation models in operational mapping applications is rare. While there are standard evaluation protocols for benchmarking, there is no clear recipe for the practical integration of foundation models into operational applications. Protocols for standardized benchmark evaluations, while invaluable for relative model comparisons under controlled conditions, do not fully capture the complexities of operational environments. 

Integrating foundation models into operational mapping applications can reveal aspects of model performance that are essential for real-world applications but are not captured by controlled benchmarks. For example:

\begin{itemize}
    \item \textbf{Operational Variability and Data Heterogeneity:} While benchmarks provide strict, standardized conditions for controlled comparisons \cite{reuel2025betterbench}, real-world deployments demand models that can flexibly adapt to dynamic environments. Variations in input data, sensor types, seasonal shifts, and different processing pipelines result in highly diverse data distributions \cite{tuia2016domain}. Consequently, a robust model must effectively handle both unpredictable operational conditions and the inherent heterogeneity of Earth Observation data.
    \item \textbf{Resource Limitations and Deployment Requirements:} Operational systems typically run on limited computational resources, often lacking access to GPU nodes, necessitating lightweight, efficient, and scalable models. Benchmark setups, however, often assume access to ample compute power and time, an assumption that rarely holds in the field. 
    \item \textbf{Deploying Something That Works:} Application developers working under operational timelines must prioritize deploying a solution that works, even if it is not the optimal solution. Due to time, budget, and resource constraints, exploring and evaluating the search space of all possible models and setups is often impractical. As a result, practitioners must make informed decisions based on limited testing, balancing performance with the practicalities of deployment. 
\end{itemize}

We propose a structured protocol for integrating foundation models in real-world mapping systems. The goal of this protocol is to provide a clear path beyond benchmarks for deploying foundation models, and as a result increase the number of foundation models tested and used in real-world systems. The insights gained from applying this protocol to foundation models in real-world systems will inform future research, steering research efforts towards the development of models that are more practical and applicable in real-world settings. We demonstrate the application of this protocol in the practical use case of global crop mapping via WorldCereal \cite{van2023worldcereal} \ref{sec:worldcereal} - an essential application for agricultural monitoring and global food security \cite{becker2023crop}.

The contributions of this paper are threefold: 
\begin{itemize} 
    \item We present a comprehensive recipe for evaluating and integrating geospatial foundation models into real-world systems, outlining key decisions and steps that guide practical deployment. 
    \item We demonstrate the application of this protocol in a concrete case study focused on global cropland and crop type mapping. 
    \item We empirically demonstrate the utility of geospatial foundation models in an operational setting, via a robust set of evaluations covering the deployment requirements.
\end{itemize}

\section{Protocol for Foundation Model Application} \label{sec:protocol}

In this section, we propose a structured protocol for the practical integration of foundation models into operational applications. This protocol provides a replicable blueprint for evaluating foundation models in an operational context as well as integrating them into practical applications, ensuring that the final deployment meets both performance and real-world usability requirements.

\begin{figure}
    \centering
    \includegraphics[width=1\linewidth]{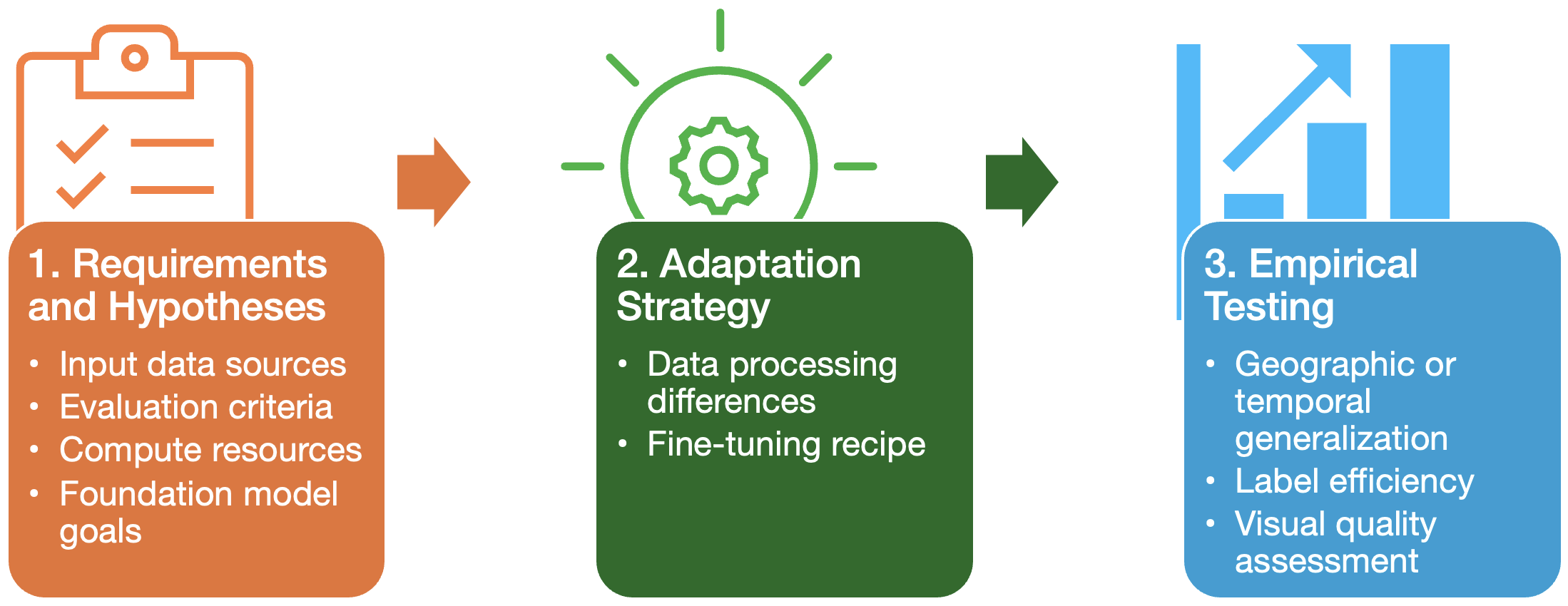}
    \caption{Visualization of the proposed three-step protocol for operationalizing foundation models in remote sensing tasks. Each step - Requirements and Hypotheses, Adaptation Strategy, and Empirical Testing - addresses specific real-world needs, ensuring that the final system balances performance with practical usability.}
    \label{fig:protocol_flow_diagram}
\end{figure}

Our protocol consists of three steps (Figure~\ref{fig:protocol_flow_diagram}): 
\begin{description} 
    \item[Step 1. Requirements and Hypotheses:]
    Clearly articulate the operational constraints and objectives. Establish which metrics or performance criteria best approximate application success.
    \item[Step 2. Adaptation Strategy:]  
    Determine what modifications will be needed to adapt the foundation model to the target application. 
    \item[Step 3. Empirical Testing:]  
    Design and execute experiments that simulate real-world scenarios, assessing the model’s performance in both standard and challenging conditions to ensure robustness. 
\end{description}

In the following subsections, we describe each step in detail and discuss how each affects the application design. In Section~\ref{sec:application}, we apply the proposed protocol to the specific task of global cropland and crop type mapping. 

\subsection{Step 1: Requirements and Hypotheses} \label{sec:requirements}

\paragraph{Requirements}
Operational deployments are often built using an existing pipeline or legacy system rather than starting from scratch. Consequently, a foundation model must conform to established data processing steps, computational budgets, and user expertise levels, rather than the application adapting to the model. Moreover, deployment contexts often require more than maximizing performance on a validation set, demanding models that also meet practical needs such as reliability, ease of integration, and efficient resource usage.

Requirements can be highly specific to each application. 
For example, latency is a key requirement for the Skylight system \cite{beukema2023satellite}, making high precision per prediction critical. On the other hand, Global Forest Watch \cite{hansen2013high} outputs an annual product, making real-time latency is less important; since Global Forest Watch's results are aggregated over regions, \textit{aggregated} accuracy is more important than the precision of individual pixel-predictions. 

Common requirements that remote sensing application developers should define when deploying foundation models include:

\begin{itemize}
\item \textbf{Evaluation criteria}: While ``traditional'' ML metrics are important, deployment contexts often introduce additional requirements on model performance. For example, when models are used to make maps, visual assessment of artifacts (e.g.,~tiling artifacts) in the maps is critical because these artifacts may not be captured in accuracy scores \cite{zvonkov2023openmapflow}. This may require the development of additional evaluation infrastructure (e.g., to visualize maps of dense predictions). 
\item{\textbf{Input data sources}}: Remote sensing practitioners have a rich variety of datasets that can be leveraged as inputs to machine learning models. For example, the SoilGrids project used over 400 products as inputs \cite{poggio2021soilgrids}. When selecting which input datasets to include, practitioners need to consider which covariates are most relevant for their task as well as the spatial and temporal resolutions of these datasets, since these will determine the resolutions of the model's predictions. These decisions directly influence the choice of which foundation model to use. For instance, if a particular task requires predictions at a higher resolution than 10 m/pixel (e.g., building damage assessment \cite{robinson2023rapid}), then foundation models trained on Sentinel-2 imagery are not ideal. If a task relies on learning temporal dynamics in input data (e.g., crop mapping \cite{garnot2019time}), the user should consider foundation models that support multi-temporal inputs.
\item \textbf{Compute resources:} Application developers do not have unlimited computational resources, and have to work within a fixed computational infrastructure and available budget for a project. Defining this requirement will influence the choice of which foundation model(s), adaptation strategies, and experiments are feasible to consider. The choice of foundation models can be especially salient in this respect because of their varying computational footprints. Comparing three models which can process pixel-timeseries in terms of their forward-pass cost for a 12-timestep timeseries yields multiply–accumulate (MAC) operations ranging from 38.37M for Presto \cite{tseng2024lightweightpretrainedtransformersremote} to 89.40M for Galileo-Nano \cite{tseng2025galileo} to 889.94M for AnySat \cite{astruc2024anysatearthobservationmodel}.
\end{itemize}

\paragraph{Hypotheses} Users typically have specific goals for applying foundation models. For example, spatial and temporal generalization in label-scarce scenarios is commonly described as a benefit of remote sensing foundation models, due to biases in the label distributions (e.g., a lack of crop type labels in Sub-Saharan Africa \cite{nakalembe2023considerations}). Users should define their hypotheses about the value of foundation models to their application and empirically validate them before deployment. 
These hypotheses impact the evaluation metrics adopted. For example, if one hypothesis is that foundation models will improve temporal or spatial generalization, the evaluation metrics must test this.

Defining these requirements and hypotheses lays the groundwork for informed decision-making throughout the deployment process. The model requirements---particularly the input data choices---can filter the choice of which foundation model to deploy, which is especially helpful when evaluating many possible models is infeasible.

\subsection{Step 2. Adaptation Strategy} \label{sec:adaptation}
Building on the specific requirements defined in Section \ref{sec:requirements}, the adaptation strategy aims to align the selected foundation model with the unique characteristics of a remote sensing application. Specific considerations for adapting foundation models to specific use cases include:

\begin{itemize}
\item \textbf{Data Processing Differences}: Distribution shifts may arise due to differences in data processing between the data a foundation model was pre-trained on versus the data it will be finetuned on. For example, models pre-trained on Sentinel-2 data are typically trained on either Level-2A data (e.g. CROMA \cite{fuller2023croma}) or Level-1C data (e.g. Presto \cite{tseng2024lightweightpretrainedtransformersremote}). If a user's application requires a specific processing level, model adaptation may be necessary. If the application has a large dataset for fine-tuning, this adaptation may be achieved during the supervised fine-tuning stage. However, if the application has few labels for the supervised fine-tuning stage, an additional self-supervised learning (SSL) stage may be needed to adapt the model to the application's data distribution before fine-tuning. We explore this in Section \ref{sec:multiclass_croptype}.

\item \textbf{To freeze or finetune?}: While foundation models consistently perform better with finetuned rather than frozen backbones \cite{tseng2025galileo,reed2023scale,cong2022satmae}, fine-tuning large models can be computationally expensive. For example, \citet{cong2022satmae} used clusters of 4 to 8 V100 GPUs to finetune the SatMAE model. Practitioners working in a compute-constrained setting need to consider this when defining the adaptation strategy they will employ. Some applications may choose to finetune the foundation model backbone, finetune a subset of layers, or use the model as a frozen feature extractor for a separate lightweight model (e.g., a multi-layer perceptron or random forest).
\end{itemize}

Defining the adaptation strategy involves determining how best to modify the foundation model so that it can process heterogeneous remote sensing data, meet task-specific demands, and operate efficiently under real-world constraints. The outcomes of these decisions will directly influence the experiments performed the evaluation of the application hypotheses (Section \ref{sec:empirical_testing}), ensuring the deployed model is both robust and practical for the operational application.

\subsection{Step 3. Empirical Testing} \label{sec:empirical_testing}

Once the application requirements and hypotheses are defined (Section \ref{sec:requirements}) and an adaptation strategy has been determined (Section \ref{sec:adaptation}), the user should perform experiments to assess the suitability of the chosen foundation model, or candidate models, for their specific application. 

This involves performing experiments that test the hypotheses defined in Section \ref{sec:requirements}. Experiment setups for testing common hypotheses include:
\begin{itemize}
\item \textbf{Geographic Generalization:}  
Design experiments that withhold data from specific regions or countries during training and then test the model on these held-out areas. This setup evaluates whether the model can generalize to regions with limited or no training labels.
\item \textbf{Temporal Generalization:}  
Simulate temporal shifts by training on historical data and testing on later periods (e.g., withholding the most recent year). This template assesses the model’s ability to capture seasonal dynamics and adapt to new temporal contexts.
\item \textbf{Label Efficiency:}  
Determine the minimum amount of local reference data required to improve performance by incrementally injecting out-of-distribution (OOD) samples into the downstream classifier. This experiment tests the model's capacity to leverage small amounts of additional data in regions with sparse labels and allows the user to estimate how many labeled examples will be needed for their task.   
\item \textbf{Visual Quality Assessment:}  
This experiment tests the prevalence of common visual artifacts in maps of model predictions. \citet{huang2018tiling} describe common tiling artifacts when conducting semantic segmentation on remote sensing data, including translational variance in patch predictions. These artifacts can be assessed by visually inspecting spatial patches from random or user-selected example regions, aiming to cover uniform ``easy'' areas to challenging corner cases. This ensures that high quantitative scores correspond to coherent, artifact-free maps.
\end{itemize}

\section{Application to Global Crop Mapping}
\label{sec:application}
In this section, we apply the protocol introduced in Section \ref{sec:protocol} to the specific case of global cropland and crop type mapping. Our goal is to demonstrate the translation of the protocol into concrete specifications and implementation for our application, addressing the unique challenges and characteristics of this task.

\subsection{WorldCereal Crop-Mapping System}\label{sec:worldcereal}
WorldCereal is a fully open, modular crop-mapping service funded by the European Space Agency.  
It combines (i) a \textbf{Reference Data Module (RDM)} for storage, harmonisation and API/GUI-based access to in-situ and map-based training data, (ii) a \textbf{Processing Module} that enables users to run either the default or their own crop classifiers over any region and season from 2017 onwards, executing the resulting crop models as openEO process graphs, and (iii) a \textbf{Visualisation \& Dissemination Module (VDM)} that allows users to browse and download default products through a web-based UI. The default product suite delivers annual cropland extent and seasonal crop-type layers for 9 major crops; users may retrain models for any custom class using the publicly available training data in their region, optionally augmented by uploading private reference data to the RDM.

Operationally, WorldCereal runs on openEO backend deployed on the Copernicus Data Space Ecosystem (CDSE). A lightweight \emph{Processing Hub} (https://hub.esa-worldcereal.org) exposes the full workflow to non-experts, whereas advanced users can drive the same processing pipelines programmatically via the openEO API and the WorldCereal Python package. Taken together, the loosely coupled architecture, openEO-based orchestration, and MIT-licensed codebase make WorldCereal FAIR and cloud-agnostic by design. And the same should apply to the model that powers the Processing Module. Additionally, that model must be well-suited for the tasks that are set to the Processing Module and deliver consistent geographical and temporal performance and flexibility that allows users to retrain local classifiers. These requirements are discussed in details and tested in the following sections. 

\subsection{Step 1: Requirements and Hypotheses} \label{sec:specific_requirements}
\paragraph{Requirements.}
Our application includes two distinct classification tasks: \textbf{binary cropland classification} (distinguishing temporary crops from all other land covers) and \textbf{multiclass crop type classification} (differentiating between multiple crop types within temporary crops). For both tasks, our goal is twofold: 1) to generate global, end-of-season 10 m/pixel prediction maps, and 2) to empower end users to produce custom maps for any spatial and temporal extent. In addition, users must be able to retrain the model and make a new map in a lightweight manner, incorporating their own labelled data and selecting relevant classes. In particular, the compute infrastructure which allows users to retrain maps \emph{does not include GPUs}, so a model which can run efficiently on a CPU is necessary.

For our application, we add the following specificities to the ``common considerations'' described in Section \ref{sec:requirements}:

\begin{itemize}
\item \textbf{Evaluation criteria:} To make accurate, global maps for the target year, we need to assess the generalization capabilities of our models for unseen years and regions. We therefore develop multiple train / val splits in addition to a naive random split. These require retraining the models on each split, but allow us to assess the generalization capabilities of the model. For all splits, we measure either per-class F1 scores, or macro F1 score of the model predictions.
\begin{itemize}
\item \textbf{Geographic evaluation}: To assess geographic generalization, we construct a ``geographic'' split where a group of countries are entirely removed from the training set, and model performance is assessed against these removed countries. Detailed description of this split for both tasks is provided in Appendix~\ref{app:spatial_split}.
\item \textbf{Temporal evaluation}: Similarly, we construct a ``temporal'' split where the latest year of labelled data (2021) is removed from the training set, and we evaluate against this removed data. 
\item \textbf{Visual quality}: We develop infrastructure to rapidly visualize ``patches'' of maps, so that we can qualitatively assess the model quality.
\end{itemize}
\item \textbf{Input data}: Our application's operational data processing pipeline uses multiple data products that have been widely adopted for crop mapping and land cover mapping applications \cite{van2023worldcereal,tseng2021cropharvest,van2021esa}:  Sentinel-1, Sentinel-2, the Copernicus Digital Elevation Model (GLO-30) \cite{CopernicusDEM}, and AgERA5 weather data \cite{boogaard2020agrometeorological}.
Each dataset is formatted as a 18-month pixel-level timeseries. 
\item \textbf{Labels Distribution and Sparsity:} For both the cropland and crop type classification tasks, labels are highly spatially imbalanced (see Appendix \ref{app:dataset_desrc}, Figures \ref{fig:cropland_geo_distr} and \ref{fig:croptype_geo_distr}). In addition, while this project aims to make maps for 2024/2025, labels were collected from before this time period (Figures \ref{fig:cropland_year_distr} and \ref{fig:croptype_year_distr}), introducing a temporal shift between the training and testing data. The highly imbalanced labels described above motivated us to investigate foundation models, which promise greater geographic and temporal generalization capabilities than fully supervised machine learning algorithms. More detailed description of the dataset is provided in Appendix \ref{app:dataset_desrc}.
\end{itemize}

Based on these requirements, we decided to focus on the Presto model~\cite{tseng2024lightweightpretrainedtransformersremote} for this mapping effort, since it is the closest match in terms of pre-training inputs and pixel timeseries format of available models. Presto is a foundation model trained on globally sampled data and specifically designed for pixel-timeseries consisting of Sentinel-1, Sentinel-2, DEM and weather data. 
In addition, since users must be able to retrain the model with their own data without a GPU, Presto was an ideal choice due to its lightweight computational cost.

\paragraph{Hypotheses.} 
Given our task-specific requirements, we aim to test the following hypotheses: 
\begin{description} 
    \item[H1:] The foundation model outperforms fully-supervised models in both binary cropland and multiclass crop type classification tasks.
    \item[H2:] The foundation model exhibits significantly improved spatial and temporal generalization capabilities. 
    \item[H3:] An additional self-supervised learning (SSL) round prior to task-specific fine-tuning enhances adaptation to shifted data distributions, particularly when additional (unlabelled) datasets are available. 
\end{description}

\subsection{Step 2. Adaptation Strategy} \label{sec:real_adaptation}

As noted above, we aimed to integrate Presto into an existing operational pipeline. Methods for processing remote sensing data differed between our pipeline and Presto's pre-training dataset construction. For example, Presto uses the least cloudy Sentinel-2 pixel within a time window and was trained on the L1C processing level, while we only keep cloud-free observations and work with the L2A processing level. We hypothesized (H3) that with few labels (in the crop type classification case), Presto may struggle to adapt to this shift in data processing. To solve this, we introduced an \emph{additional} SSL step, allowing Presto to adapt to our data processing methods fine-tuning it using task-specific labels.

For both the cropland and crop type tasks, we evaluated:

\begin{itemize}
\item \textbf{Finetuned Presto:} We finetuned Presto on the labelled training set, and evaluated it on the validation set.
\item \textbf{SSL + Finetuned Presto} We \emph{first} applied an SSL step to Presto, to adapt it to our data processing methods. We used the larger cropland dataset for SSL. We then finetuned Presto on the labelled training set and evaluated it on the validation set.
\end{itemize}

\subsection{Step 3. Empirical Testing} \label{sec:real_evals}

We ran two sets of experiments: one to evaluate Presto's suitability for cropland mapping,  and one to evaluate its suitability for crop type mapping. For both sets of experiments, we defined three splits (as discussed in Section \ref{sec:specific_requirements}):

\begin{itemize}
\item \textbf{Geographic split}: We held out specific countries from the training sets and evaluated on them. For the cropland experiments, we held out Spain, Nigeria, Latvia, Tanzania, Ethiopia, Argentina. For crop type, we held out Spain, Latvia, Austria, Brazil, Tanzania, Ethiopia, Madagascar, Mozambique, Morocco, Indonesia. We explain the motivation for these selections in Appendix \ref{app:spatial_split}.
\item \textbf{Temporal split}: We held out all samples from 2021 from the training set, and evaluated the models on them for both the cropland and crop type tasks.
\item \textbf{Random split}: A random 80/20\% train/validation split to evaluate model performance on in-distribution data.
\end{itemize}

\subsubsection{Binary Cropland Classification} \label{sec:results_crop_noncrop}
In Section \ref{sec:specific_requirements}, we hypothesized that using a foundation model would outperform a fully-supervised model (H1) and improve spatial and temporal generalization capabilities (H2). 
To compare Presto's performance to a fully-supervised model, we chose an \textbf{existing deployed model}, the WorldCereal cropland classifier \cite{van2023worldcereal}, as a baseline (``Deployed Baseline''). This baseline is a CatBoost model, trained separately for each continent, on expert-defined features computed from the pixel-timeseries dataset at a 10-day temporal resolution. We also assessed the performance of the same CatBoost model on the raw pixel-timeseries input (``Unprocessed CatBoost'').

To isolate the effect of Presto's self-supervised pre-training vs.~the model's transformer-based architecture, we evaluated the performance of a randomly initialized, finetuned Presto model (``Finetuned Presto-Rnd'').

\paragraph{Results.}

We report overall results in Table \ref{tab:binary_results}. These overall results strongly support hypotheses 1 and 2; the pre-trained Presto models are the best performing models, with or without the additional SSL step. We found that the SSL step did not improve performance compared to just finetuning; we hypothesize that this is because the SSL step uses the same data as the finetuning step and so (for the cropland task) does not provide an additional learning signal to the model. The randomly initialized Presto architecture performs worst, showing that Presto's pre-training significantly contributes to its performance for our application.

In addition to the overall metrics, we evaluated model performance on a per-country basis using a geographic split. Table~\ref{tab:cropland_spatial} shows Crop F1 scores for each model across all held-out countries. These spatial results confirm the overall trends: the pre-trained Presto models consistently outperform the baseline and the randomly initialized variant across diverse regions.

\begin{figure}
    \centering
    \includegraphics[width=1\linewidth]{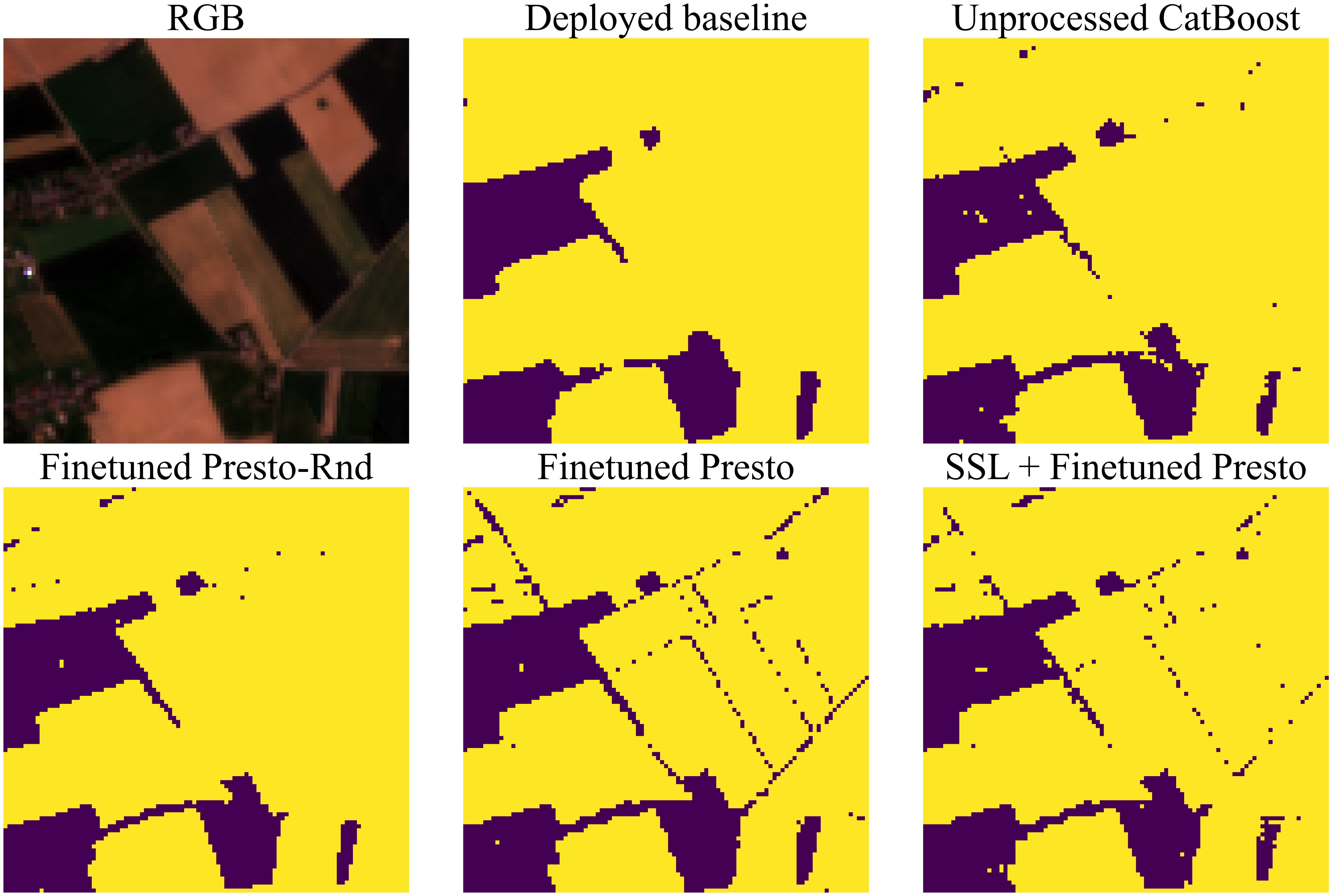}
    \caption{Qualitative comparison of cropland models trained using the ``Random'' split. Presto models show cleaner field boundaries. Patch location: Belgium.
    }
    \label{fig:cropland_sanity_check_patch}
\end{figure}

Figure \ref{fig:cropland_sanity_check_patch} visualizes a spatial patch of each model's predictions. The pre-trained Presto models show better separation between field boundaries than other models. Additional patches are shown in Appendix \ref{app:more_spatial_patches} (Figures \ref{fig:cropland_argentina_patch} and \ref{fig:cropland_usa_patch}), confirming the general findings and showing better detalization of the Finetuned Presto model.

For the cropland classification tasks, these results confirm our hypotheses: (H1) Presto outperforms a fully supervised approach, (H2) in particular demonstrating strong temporal and geographic generalization. However, we find that the SSL step (H3) is an unnecessary step here.

\begin{table}[h!]
    \centering
    \resizebox{0.5\textwidth}{!}{%
    \large
    \begin{tabular}{lccc}
         \toprule & \textbf{Random} & \textbf{Geographic} & \textbf{Temporal} \\
         \midrule Deployed baseline & 0.856 & 0.810 & 0.830 \\ 
         Unprocessed CatBoost & 0.828 & 0.777 & 0.874 \\ 
         Finetuned Presto-Rnd & 0.810 & 0.705 & 0.806 \\
         \midrule
         Finetuned Presto & \textbf{0.861} & \textbf{0.829} & \textbf{0.886} \\ 
         SSL + Finetuned Presto & \textbf{0.861} & 0.826 & 0.884 \\
         \bottomrule 
    \end{tabular}
    }
    \caption{F1 scores for the binary cropland classification task, measured across three splits (described in Section \ref{sec:real_evals}).
    } 
    \label{tab:binary_results}
\end{table}

\begin{table}[h!]
\centering
\resizebox{0.5\textwidth}{!}{%
\begin{tabular}{lcccccc}
\toprule
\textbf{Model} & \textbf{Argentina} & \textbf{Ethiopia} & \textbf{Latvia} & \textbf{Nigeria} & \textbf{Spain} & \textbf{Tanzania} \\
\midrule
Deployed baseline                           & 0.745 & 0.692 & 0.851 & 0.857 & 0.730 & 0.403 \\
Unprocessed CatBoost            & 0.896   & 0.639   & 0.866   & 0.705   & 0.738   & 0.258   \\
Finetuned Presto-Rnd         & 0.824 & 0.511 & 0.749 & 0.785 & 0.650 & 0.300 \\
\midrule
Finetuned Presto                  & 0.910 & 0.683 & \textbf{0.882} & \textbf{0.872} & 0.748 & \textbf{0.442} \\
SSL + Finetuned Presto     & \textbf{0.912} & \textbf{0.732} & 0.881 & 0.857 & \textbf{0.749} & 0.416 \\
\bottomrule
\end{tabular}%
}
\caption{Comparison of per-country Crop F1 Scores in the Geographic Split, with models as rows and countries as columns. Results demonstrate that Presto-based models outperform the baseline across various regions.}
\label{tab:cropland_spatial}
\end{table}

\subsubsection{Multiclass Crop type Classification} \label{sec:multiclass_croptype}

This task classifies a pixel as one of 8 crop types: maize, wheat, barley, soybean, millet/sorghum, sunflower, rapeseed, or other.
As in the binary cropland task (Section \ref{sec:results_crop_noncrop}), we hypothesized that a foundation model would outperform a fully supervised model (H1) and demonstrate improved spatial and temporal generalization capabilities (H2). 
Since our dataset of crop type labels is significantly smaller than the binary cropland dataset (Appendix \ref{app:dataset_desrc}), we hypothesized that performing the additional SSL pre-training step using the binary cropland samples before supervised fine-tuning on the crop type labels would allow the model to better adapt to the distribution shift introduced by different data processing algorithms (H3). 

Since there is no pre-existing deployed baseline for the crop type task, we used only the CatBoost model directly trained on the pixel-timeseries as the supervised baseline (``Unprocessed CatBoost''). We also evaluated the randomly initialized, finetuned Presto model (``Finetuned Presto-Rnd'').

\paragraph{Results.}

Table \ref{tab:croptype_macro_f1} shows overall results for the crop type classification task. As in the cropland case, these results strongly support hypotheses 1 and 2, that pre-trained Presto models outperform supervised baselines across all splits, and do not support hypothesis 3---additional SSL pre-training does not substantially improve performance. This suggests that the different data processing levels do not affect the performance of the foundation model after supervised fine-tuning, even with a small labeled dataset.

We find that Presto yields especially large increases in performance for under-represented crop types (Table \ref{tab:croptype_per_crop_f1_with_support_ordered}). For example, Presto yields a 0.13 increase in F1 score (in both the finetuned and SSL + finetuned settings) for the ``millet / sorghum'' class, which represents only 1.3\% of the crop type labels (Appendix, Figure \ref{fig:croptype_classes_distr}).

Moreover, the detailed per-country analysis on the Geographic Split (Table~\ref{tab:croptype_spatial}) confirms that the pre-trained Presto models generalize effectively to unseen regions. In diverse countries, both the Finetuned and SSL + Finetuned variants consistently achieve higher F1 scores than the baseline, with particularly notable improvements in challenging regions.

\begin{figure}
    \centering
    \includegraphics[width=1\linewidth]{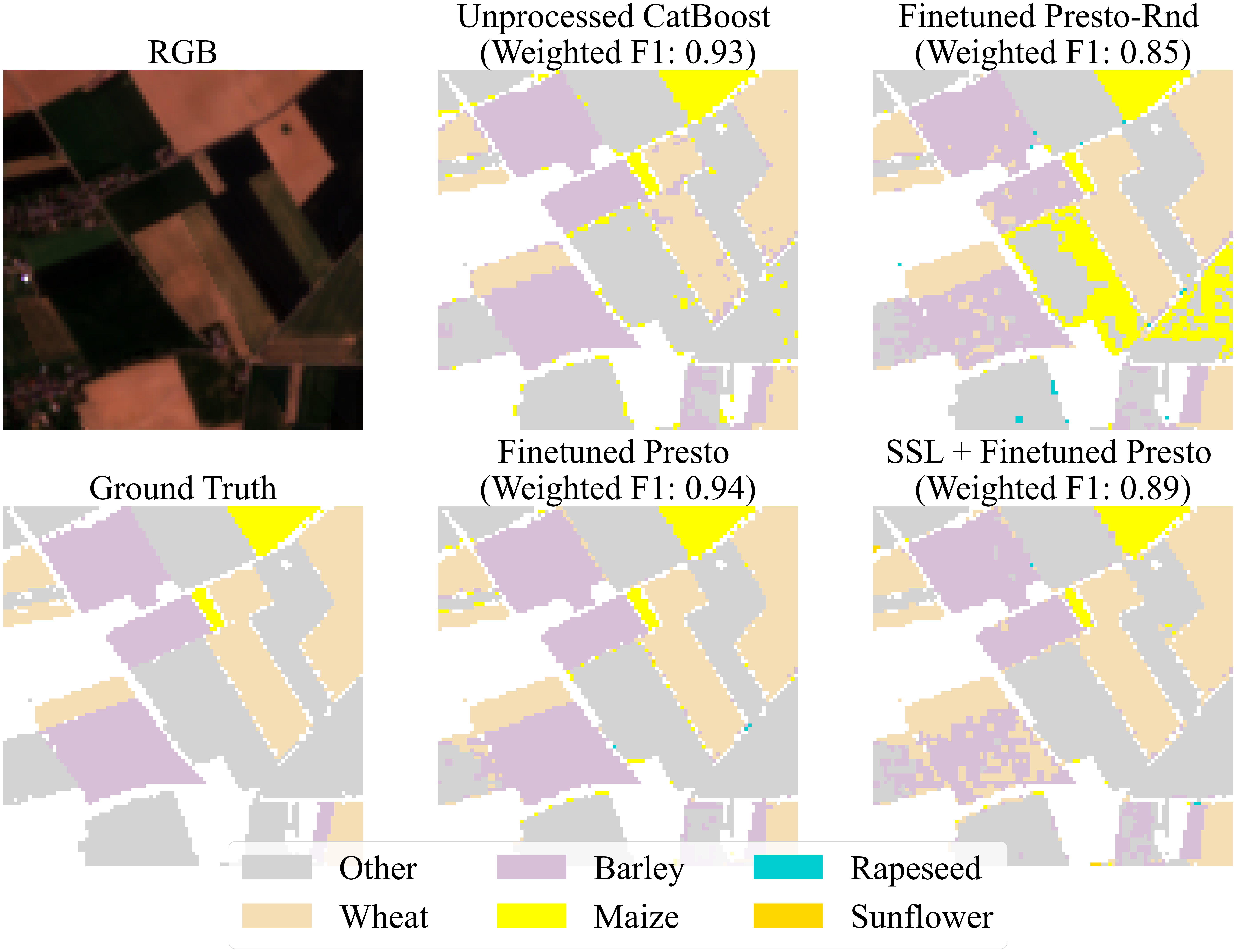}
    \caption{A qualitative assessment of the crop type results. All models were trained using the ``Random'' split before generating these patches. The pre-trained Presto models provide qualitatively good outputs, with less noise and higher overall F1 score.}
    \label{fig:croptype_sanity_check_patch}
\end{figure}

Visual inspection of spatial patches (Figure~\ref{fig:croptype_sanity_check_patch}) reinforces these findings. The Finetuned Presto model produces the cleanest outputs - with well-defined field boundaries closely matching ground truth - whereas the SSL + Finetuned model, although competitive overall, exhibits some localized noise. Additional qualitative assessments in the Appendix \ref{app:more_spatial_patches}, including a USA patch highlighting a prominent maize-soybean pattern \ref{fig:croptype_usa_patch} and an Argentinian patch where barley is expected \ref{fig:croptype_argentina_patch}, provide further insights into model performance.

Together, these results demonstrate that pre-trained Presto models deliver robust performance and strong generalization in multiclass crop type mapping, even under geographically diverse and challenging conditions.

\begin{table}[h!]
    \centering
    \resizebox{0.5\textwidth}{!}{%
    \large
    \begin{tabular}{lccc}
         \toprule & \textbf{Random} & \textbf{Geographic} & \textbf{Temporal} \\
         Unprocessed CatBoost & 0.728 & 0.563 & 0.649 \\ 
         Finetuned Presto-Rnd & 0.782 & 0.620 & 0.646 \\
         \midrule
         Finetuned Presto & 0.809 & \textbf{0.650} & \textbf{0.686} \\ 
         SSL + Finetuned Presto & \textbf{0.820} & 0.645 & 0.674 \\
         \bottomrule 
    \end{tabular}
    }
    \caption{F1 scores for the multiclass crop type classification task, measured across three splits (described in Section \ref{sec:real_evals}).
    } 
    \label{tab:croptype_macro_f1}
\end{table}

\begin{table*}[h!]
\centering
\resizebox{\textwidth}{!}{%
\begin{tabular}{l c c | c c}
\toprule
\textbf{Country (Val Samples)} & \textbf{Unprocessed CatBoost} & \textbf{Random} & \textbf{Finetuned Presto} & \textbf{SSL + Finetuned Presto} \\
\midrule
Austria (33.8K) & 0.519 & 0.558 (+0.039) & 0.605 (+0.086) & \cellcolor{green!25}\textbf{0.611 (+0.092)} \\
Spain (21.2K)   & 0.400 & 0.481 (+0.081) & \cellcolor{green!25}\textbf{0.516 (+0.116)} & 0.506 (+0.106) \\
Brazil (0.9K)   & 0.563 & 0.675 (+0.112) & 0.745 (+0.182) & \cellcolor{green!45}\textbf{0.756 (+0.193)} \\
Italy (0.6K)    & 0.614 & 0.584 (–0.030) & 0.623 (+0.009) & \cellcolor{green!25}\textbf{0.647 (+0.033)} \\
Madagascar (0.5K)& 0.432 & 0.487 (+0.055) & 0.479 (+0.047) & \cellcolor{green!25}\textbf{0.518 (+0.086)} \\
Mozambique (0.4K)& 0.397 & 0.287 (–0.110) & \cellcolor{green!25}\textbf{0.437 (+0.040)} & 0.396 (–0.001) \\
Ethiopia (0.2K)  & 0.492 & 0.541 (+0.049) & 0.559 (+0.067) & \cellcolor{green!25}\textbf{0.631 (+0.139)} \\
Greece (0.2K)    & 0.581 & 0.606 (+0.025) & \cellcolor{green!25}\textbf{0.663 (+0.082)} & 0.645 (+0.064) \\
Morocco (0.2K)   & 0.228 &  0.270 (+0.042) & \cellcolor{green!40}\textbf{0.384 (+0.156)} & 0.326 (+0.098) \\
\bottomrule
\end{tabular}%
}
\caption{Per-country macro F1 scores for multiclass crop type classification (\textbf{Geographic Split}). We compare the Unprocessed CatBoost baseline with three Presto-based approaches. For the columns \textbf{Random}, \textbf{Finetuned Presto}, and \textbf{SSL + Finetuned Presto}, values in parentheses represent the change relative to the baseline. Cells containing the best result in each row are highlighted in green, with intensity proportional to the magnitude of the gain. Since this is a geographic split, the number of training samples for each country is 0.}
\label{tab:croptype_spatial}
\end{table*}

\begin{table*}[h!]
\centering
\resizebox{\textwidth}{!}{%
\begin{tabular}{l r r | r r}
\toprule
\textbf{Class (Train/Val samples)} & \textbf{Unprocessed CatBoost} & \textbf{Random} & \textbf{Finetuned Presto} & \textbf{SSL + Finetuned Presto} \\
\midrule
Maize (63.2K/4.3K)          & 0.878 & 0.892 (+0.014) & 0.903 (+0.025) & \colorbox{green!25}{\strut \textbf{0.910 (+0.032)}} \\
Wheat (51.4K/4.4K)          & 0.774 & 0.791 (+0.017) & 0.814 (+0.040) & \colorbox{green!40}{\strut \textbf{0.816 (+0.042)}} \\
Other Crop (46.8K/3.2K)     & 0.698 & 0.727 (+0.029) & 0.756 (+0.058) & \colorbox{green!60}{\strut \textbf{0.769 (+0.071)}} \\
Barley (27.5K/2.5K)         & 0.679 & 0.694 (+0.015) & 0.728 (+0.049) & \colorbox{green!50}{\strut \textbf{0.729 (+0.050)}} \\
Sunflower (21.7K/1.8K)      & 0.890 & 0.911 (+0.021) & \colorbox{green!25}{\strut \textbf{0.923 (+0.033)}} & 0.919 (+0.029) \\
Rapeseed (18.1K/1.7K)       & 0.916 & 0.934 (+0.018) & 0.939 (+0.023) & \colorbox{green!25}{\strut \textbf{0.946 (+0.030)}} \\
Soybeans (16.5K/1.5K)       & 0.853 & 0.878 (+0.025) & 0.883 (+0.030) & \colorbox{green!50}{\strut \textbf{0.908 (+0.055)}} \\
Millet / Sorghum (5.2K/0.1K) & 0.415 & 0.429 (+0.014) & 0.530 (+0.115) & \colorbox{green!85}{\textbf{0.563 (+0.148)}} \\
\midrule
Macro F1 (254K/19.6K)       & 0.728 & 0.782 (+0.054) & 0.809 (+0.081) & \colorbox{green!70}{\strut \textbf{0.820 (+0.092)}} \\
\bottomrule
\end{tabular}%
}
\caption{Per-crop F1 scores for multiclass crop type classification (\textbf{Random Split}), with support values for Train/Val indicated in parentheses after each class name. The table compares the Unprocessed CatBoost baseline with three Presto-based approaches. In the Random, Finetuned Presto, and SSL + Finetuned Presto columns, values in parentheses represent the change relative to the baseline. Cells containing the best result in each row are highlighted in green, with intensity proportional to the magnitude of the gain.}
\label{tab:croptype_per_crop_f1_with_support_ordered}
\end{table*}

\subsection{Application Summary} 
In this section, we comprehensively evaluated our foundation model–based approach for global cropland and crop type mapping. Our results strongly support H1 and H2: the pre-trained Presto model consistently outperforms conventional baselines and exhibits robust spatial and temporal generalization. In the multiclass crop type task, the additional SSL round did not yield appreciable improvements, thereby disproving H3 and suggesting that anticipated benefits from addressing data-processing shifts are limited for Presto. Together, these results validate our protocol and underscore the practical potential of foundation models for operational remote sensing applications.

\section{Lessons Learned}

The application of our foundation model deployment protocol to global cropland mapping revealed several key insights that we feel are useful to highlight for future practitioners integrating foundation models into operational remote sensing applications.

\begin{itemize}
\item \textbf{Task-specific alignment is crucial:} Pre-trained foundation models offer a robust starting point, yet further fine-tuning on domain-specific data often yields only modest gains in performance and generalization. This highlights the importance of selecting a model that already closely aligns with the target data characteristics.
\item \textbf{Pre-trained models distill useful patterns:} Despite differences between pre-training and target data distributions, foundation models capture transferable representations that outperform models trained solely on limited, task-specific data.
\item \textbf{Evaluating beyond benchmarks:} Standard benchmarks do not capture the full complexity of real-world deployment. Our work demonstrates the need for comprehensive evaluations - addressing geographic and temporal generalization as well as qualitative map quality—to assess a model's operational utility.
\item \textbf{Computational efficiency matters:} In resource-constrained environments, lightweight models are essential to enable experimentation, iteration, and efficient inference for large-scale maps. Balancing performance with computational cost is critical.
\item \textbf{Compatibility with existing systems eases integration:} Successful deployment hinges on seamless integration into existing workflows and data pipelines. Our experience underscores the need for models that not only achieve high accuracy but also conform to practical constraints such as legacy system compatibility and ease of maintenance.
\end{itemize}

\section{Conclusion}
We present a generic protocol for integrating foundation models into operational remote sensing applications and demonstrate the application of this protocol to the specific application of global cropland and crop type mapping. 

Our results show that leveraging a pre-trained foundation model significantly outperforms conventional approaches, delivering robust spatial and temporal generalization in both binary and multiclass tasks. Pre-training is crucial for capturing the diverse features of heterogeneous remote sensing data, while our experiments show that additional adaptation steps, such as extra self-supervised learning rounds, do not yield appreciable gains for crop type mapping, underscoring the task-dependent nature of such enhancements.

Overall, our three-step protocol provides a replicable blueprint that bridges the gap between controlled benchmark evaluations and the practical challenges of real-world deployment. This balanced approach, which incorporates domain-specific adaptations and addresses resource constraints, lays a solid foundation for future efforts to operationalize foundation models across a wide range of remote sensing applications.

\section*{Acknowledgements}

This project was supported in part through the NASA Harvest Grant \#80NSSC23M0032, the European Space Agency  (grant no. 4000130569/20/I-NB), the Canada CIFAR AI Chairs program and the NSERC-CREATE LEADS program.

\bibliography{example_paper}
\bibliographystyle{icml2025}

\appendix

\section{Dataset Description} \label{app:dataset_desrc}
\begin{figure}
    \centering
    \includegraphics[width=1\linewidth]{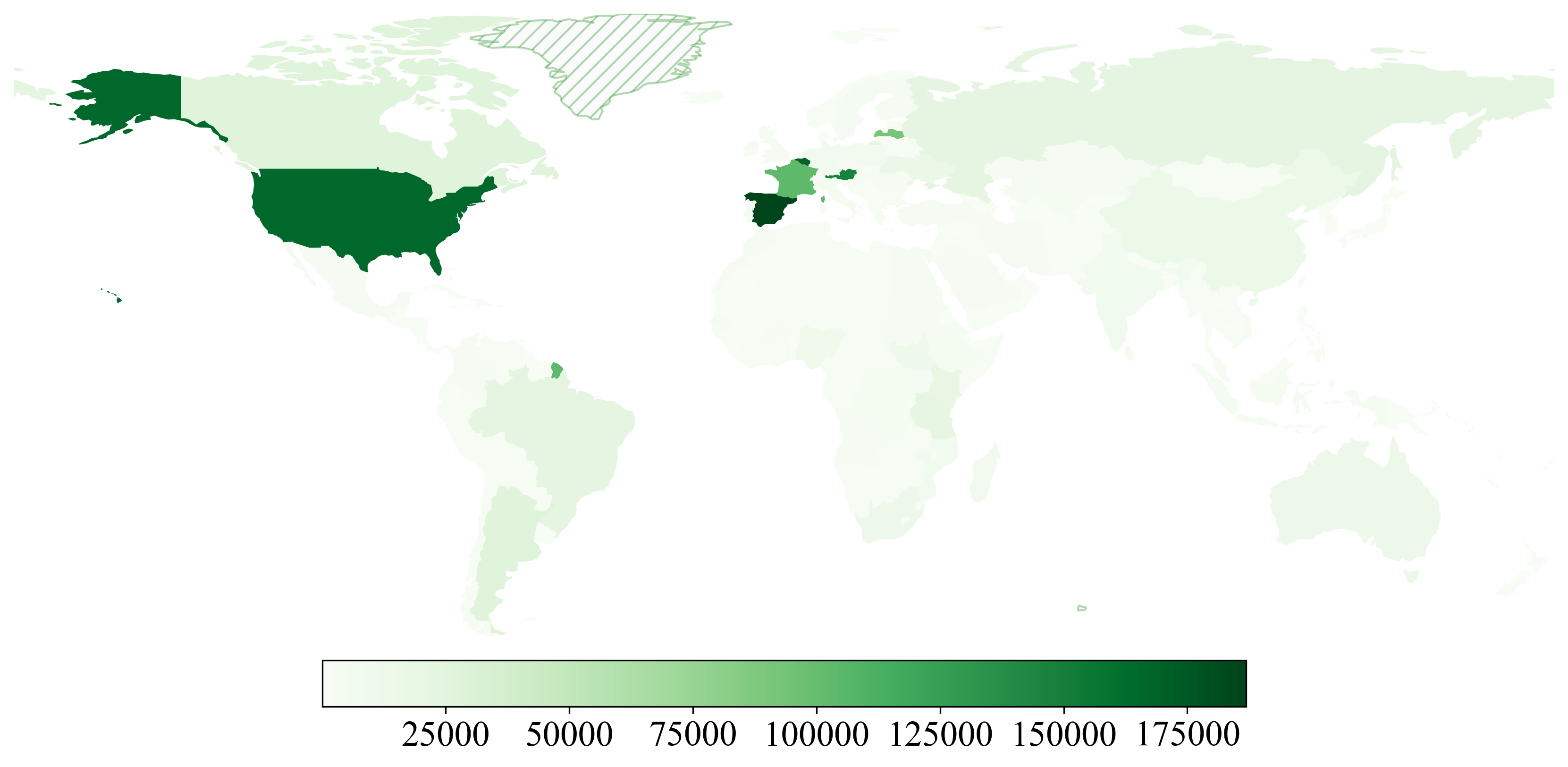}
    \caption{Geographic Distribution of Cropland Labels}
    \label{fig:cropland_geo_distr}
\end{figure}

\begin{figure}
    \centering
    \includegraphics[width=0.5\linewidth]{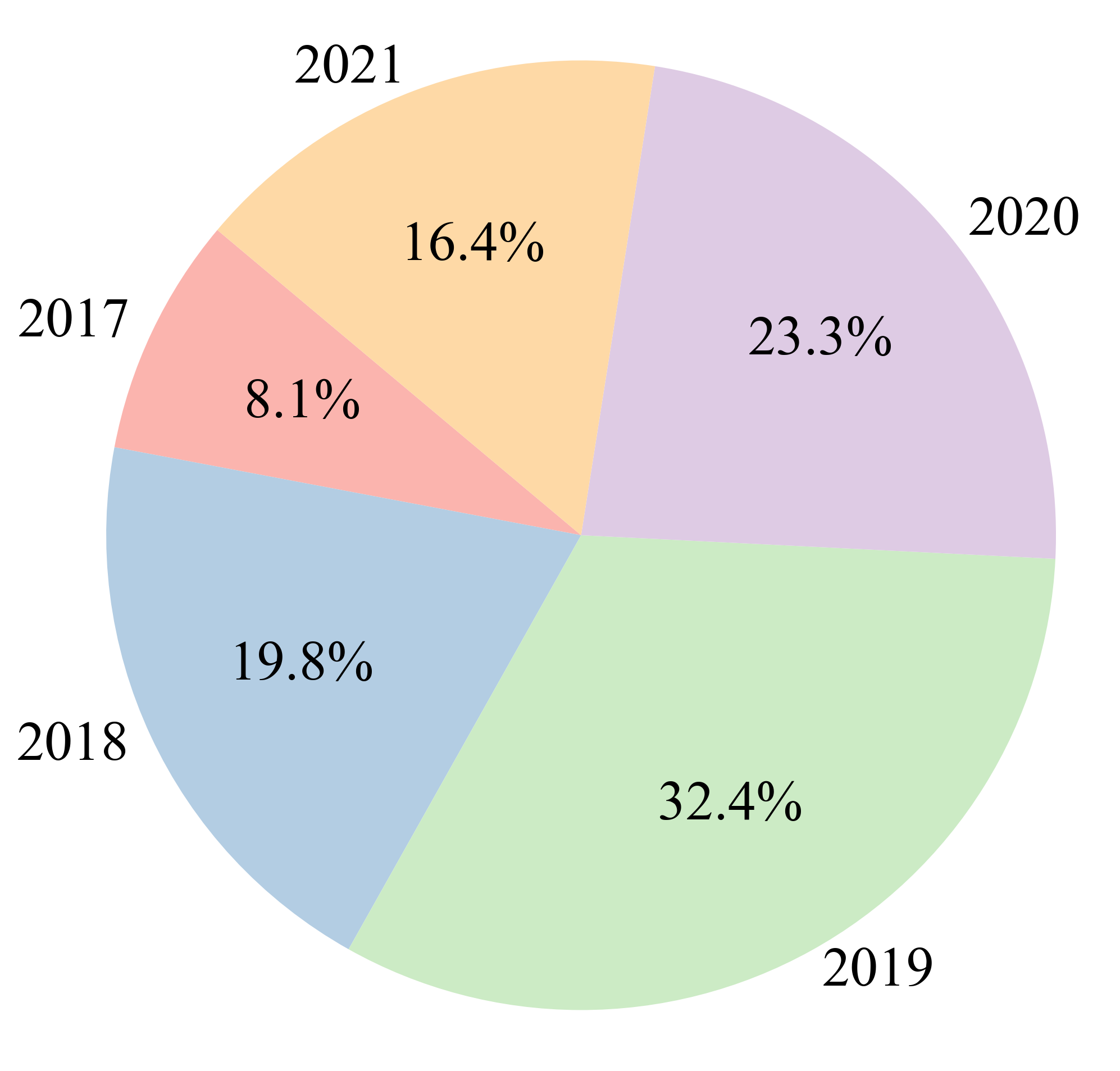}
    \caption{Temporal Distribution of Cropland Labels}
    \label{fig:cropland_year_distr}
\end{figure}

The \textbf{cropland} dataset comprises sampled point data with the following key characteristics:
\begin{itemize}
    \item \textbf{Total Samples:} Approximately 1.3 million points, with 26\% labeled as cropland.
    \item \textbf{Geographic Coverage:} Data from 176 countries; notably, the USA, Spain, and Belgium account for 40\% of the dataset, indicating significant spatial imbalance (see Figure~\ref{fig:cropland_geo_distr}).
    \item \textbf{Temporal Coverage:} Spanning five years (2017--2021) with a relatively even distribution (see Figure~\ref{fig:cropland_year_distr}).
    \item \textbf{Data Sources:} Aggregated from 121 sources, including prominent contributions from the USDA Crop Data Layer \cite{USDA_NASS_CDL2021}, the LUCAS Copernicus 2018 dataset \cite{dAndrimont2021}, and various European farmers’ declaration datasets.
\end{itemize}

\begin{figure}
    \centering
    \includegraphics[width=1\linewidth]{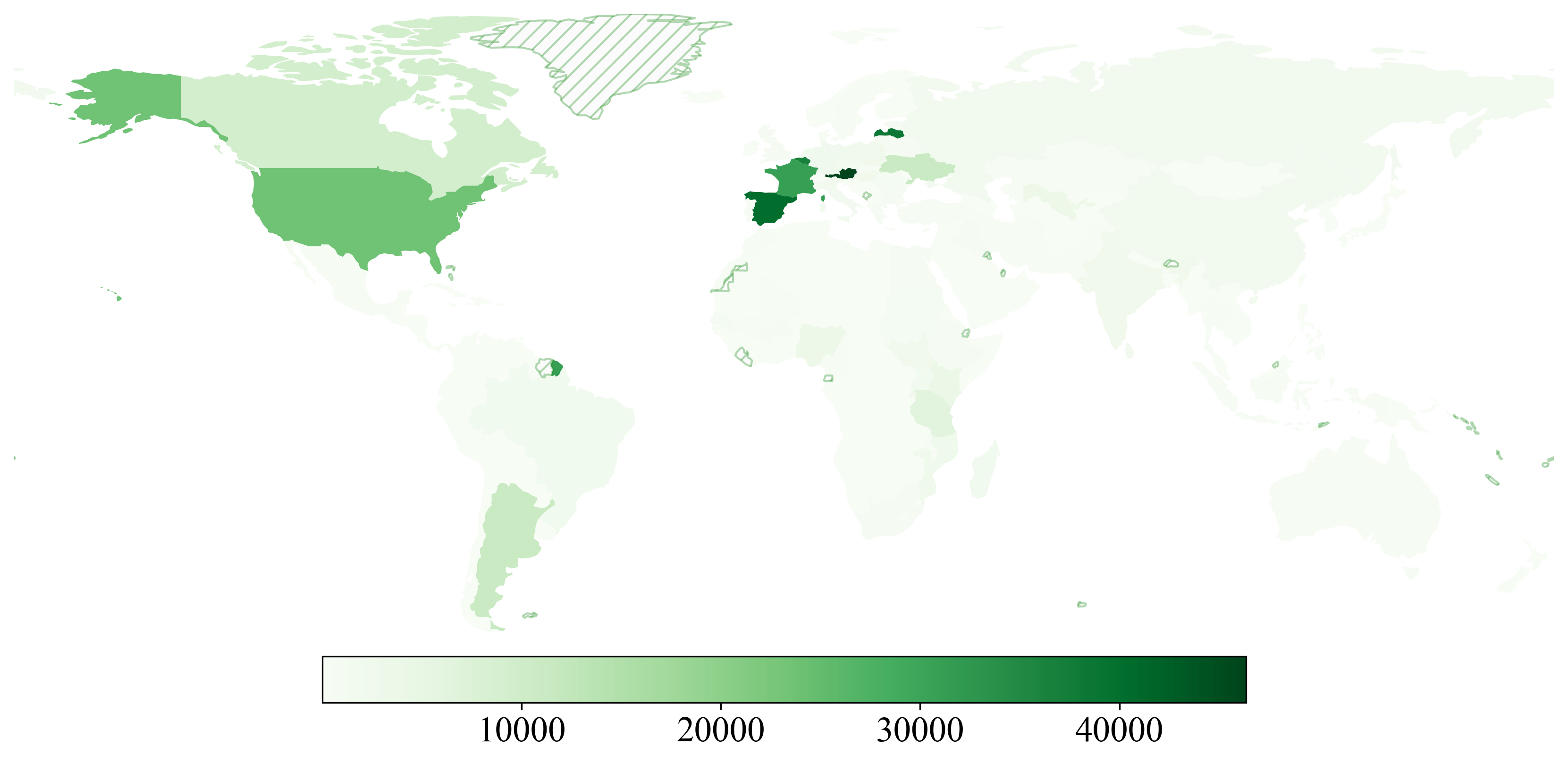}
    \caption{Geographic Distribution of Crop type Labels}
    \label{fig:croptype_geo_distr}
\end{figure}

\begin{figure}
    \centering
    \includegraphics[width=0.5\linewidth]{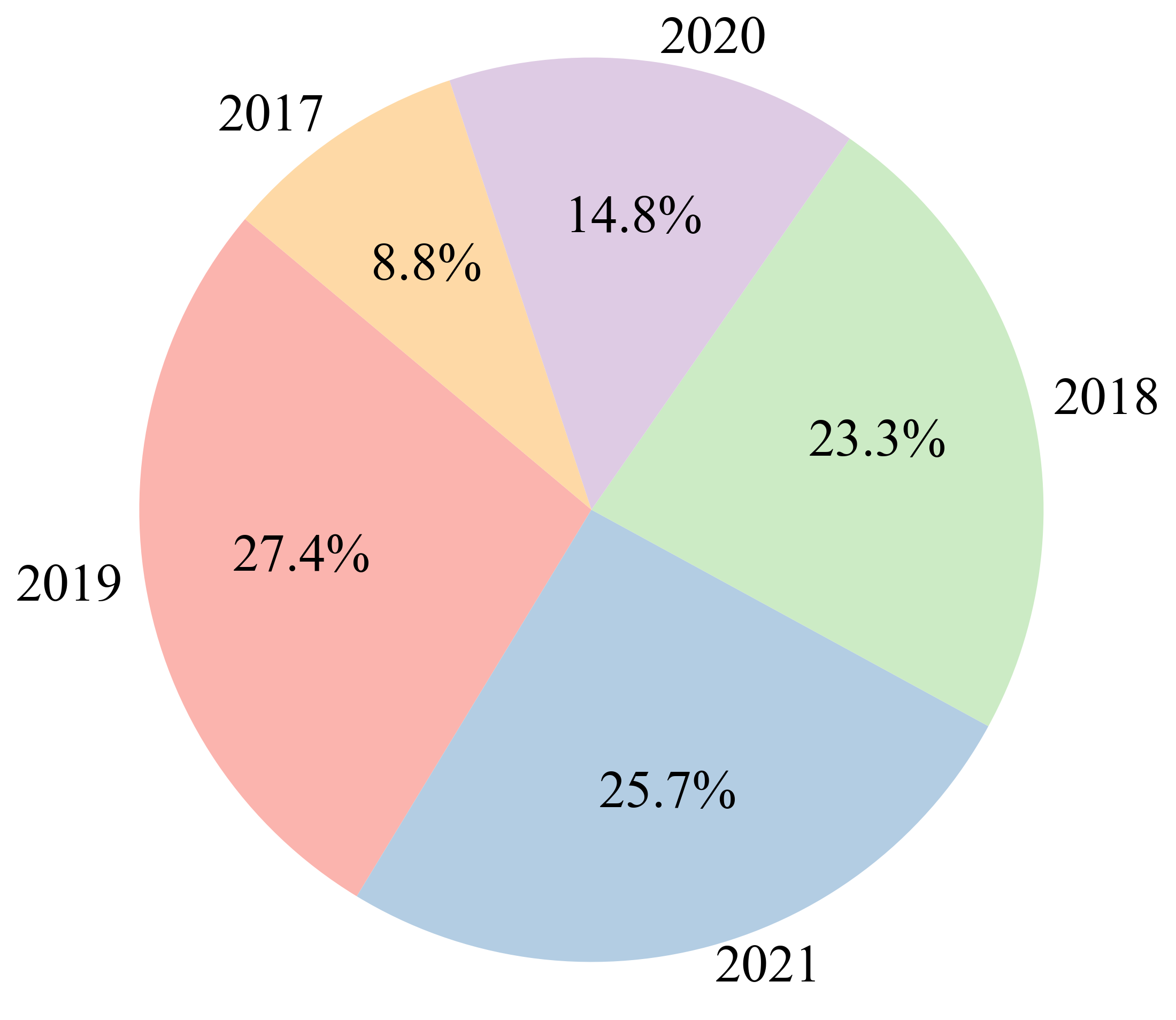}
    \caption{Temporal Distribution of Crop type Labels}
    \label{fig:croptype_year_distr}
\end{figure}

For the \textbf{crop type} task, we use a subset of the above dataset with detailed crop type labels. In our application, we consider the following classes:
\begin{center}
\texttt{maize, wheat, barley, soybeans, millet/sorghum, sunflower, rapeseed}
\end{center}
All other classes are grouped into an \texttt{"other\_crop"} category. Additional characteristics of the crop type dataset are:
\begin{itemize}
    \item \textbf{Total Samples:} 255,000 points with crop type labels.
    \item \textbf{Geographic Coverage:} Sparser and more skewed than the cropland dataset, with several European countries and the USA dominating (see Figure~\ref{fig:croptype_geo_distr}).
    \item \textbf{Temporal Coverage:} An even five-year distribution, similar to the cropland data (see Figure~\ref{fig:croptype_year_distr}).
\end{itemize}

\begin{figure}
    \centering
    \includegraphics[width=1\linewidth]{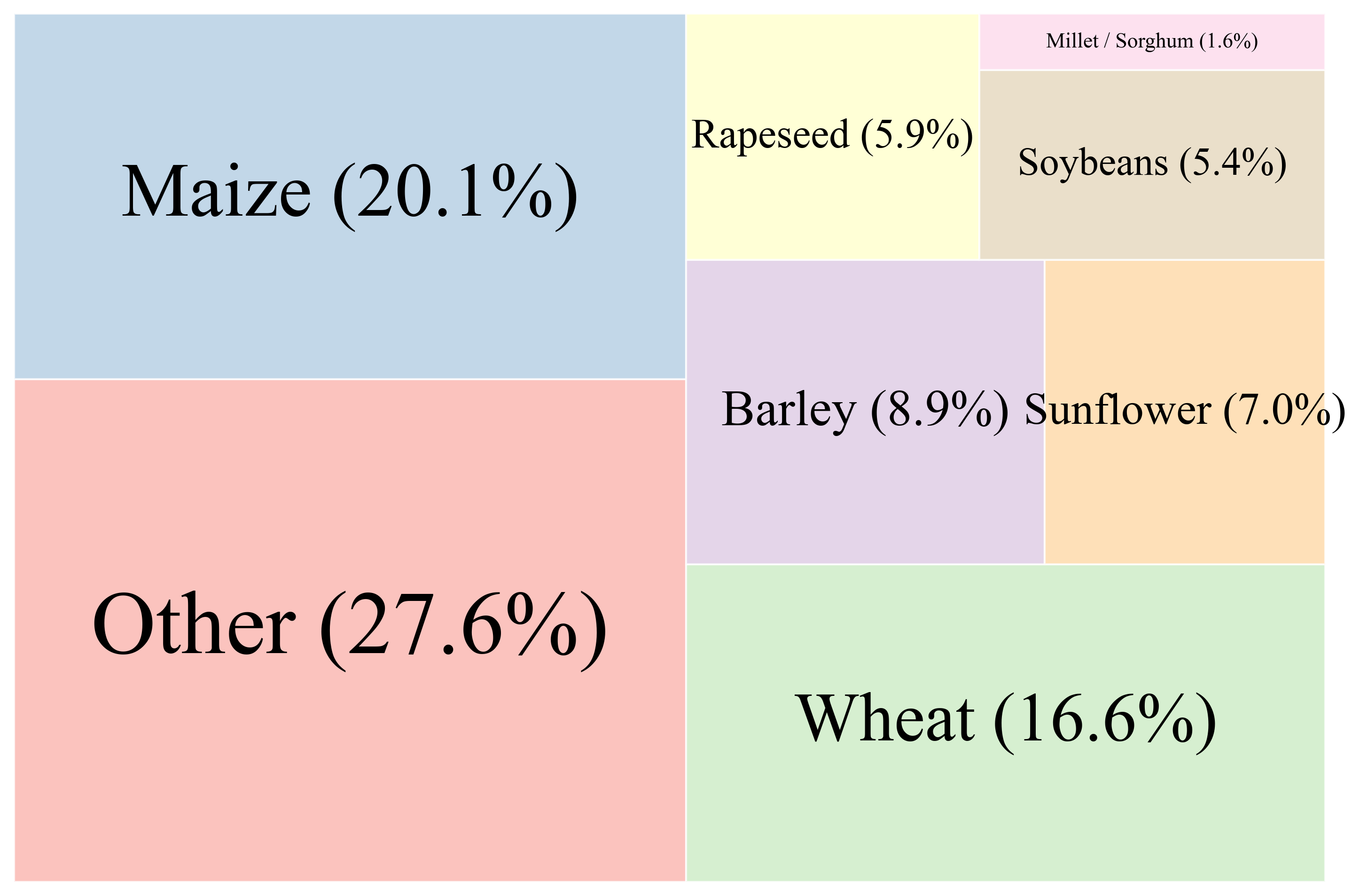}
    \caption{Crop Type Classes Distribution}
    \label{fig:croptype_classes_distr}
\end{figure}

\section{Spatial Split Description} \label{app:spatial_split}
To assess geographic generalization, we defined a spatial split by excluding a selected set of countries from training and validation, then evaluating model performance solely on these held‐out regions. The selection was based on geographic variability, sample size, and label quality, ensuring representation of both straightforward and challenging cases.
For the \textbf{cropland} task, we excluded the following countries:
\begin{itemize}
    \item \textbf{Spain:} A large European dataset (66K samples over four years) with moderate quality, complex seasonality, and diverse crop distributions.
    \item \textbf{Nigeria:} A smaller dataset (approx. 5K samples, about half labeled as cropland) spanning 2019–2020 from different sources, where cropland is harder to define.
    \item \textbf{Latvia:} A very large, high-quality dataset (69K points, including 32.2K cropland points) spanning three years, with easily detectable cropland patterns.
    \item \textbf{Tanzania:} A dataset of 6.7K points (1.4K cropland) from two sources (2019–2020) where cropland boundaries are less distinct.
    \item \textbf{Ethiopia:} A relatively small dataset (2K points with 0.5K cropland) from 2018 and 2020, expected to be challenging due to ambiguous definitions.
    \item \textbf{Argentina:} Representing Latin America with 15K samples (9.7K cropland) from several high-quality sources over three years, offering a balanced test case.
\end{itemize}
For the \textbf{crop type task}, we selected a geographically diverse and challenging set of countries based on expert assessments and prior results, aligning our selection as closely as possible with that of the cropland task:
\begin{itemize}
    \item \textbf{Europe:} Spain, Latvia, and Austria were selected, with Austria added to capture a broader range of crop types not present in Spain and Latvia alone, and supplemented by several smaller-label European countries for comprehensive regional representation.
    \item \textbf{South America:} Brazil was chosen over Argentina, as its smaller dataset (less than 1K samples) provides a more diverse regional representation.
    \item \textbf{Africa:} Tanzania, Ethiopia, Mozambique, Madagascar, and Morocco were included to capture a broad spectrum of crop types and label challenges.
    \item \textbf{Asia:} Indonesia was added to represent the region, despite being a small croptype-only dataset.
\end{itemize}

This spatial split enables a rigorous evaluation of the model’s geographic generalization by testing its performance in regions with varying label density, quality, and complexity, thereby providing a realistic measure of its operational effectiveness.

\section{Additional Spatial Patches}
\label{app:more_spatial_patches}

\begin{figure}
    \centering
    \includegraphics[width=1\linewidth]{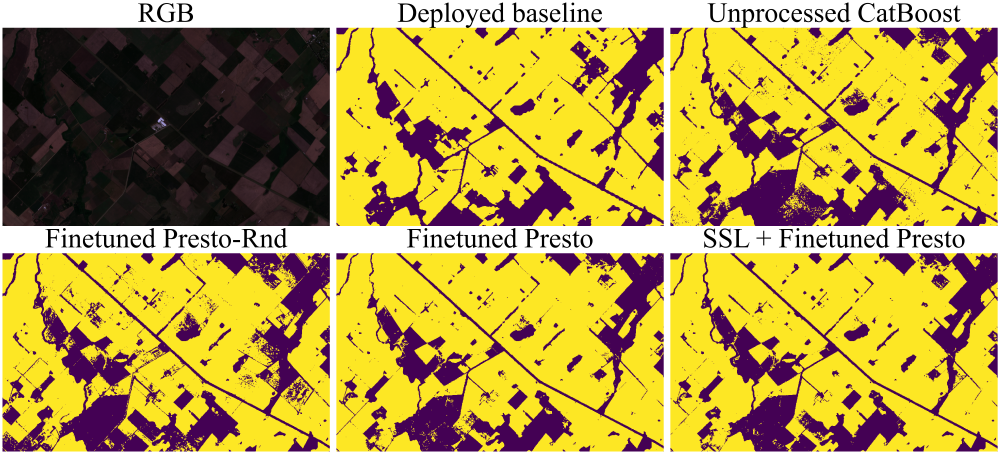}
    \caption{Qualitative comparison of cropland models trained using the ``Random'' split. Presto models show cleaner field boundaries. Patch location: Argentina.
    }
    \label{fig:cropland_argentina_patch}
\end{figure}

\begin{figure}
    \centering
    \includegraphics[width=1\linewidth]{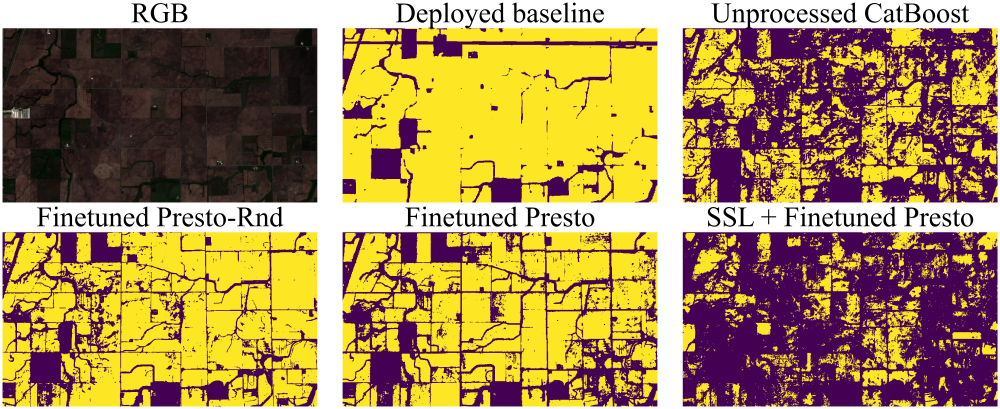}
    \caption{Qualitative comparison of cropland models trained using the ``Random'' split. Finetuned Presto model shows cleaner field boundaries, while SSL + Finetuned model shows a much worse result comparable to the Unprocessed CatBoost model.  Patch location: USA.
    }
    \label{fig:cropland_usa_patch}
\end{figure}

\begin{figure}
    \centering
    \includegraphics[width=1\linewidth]{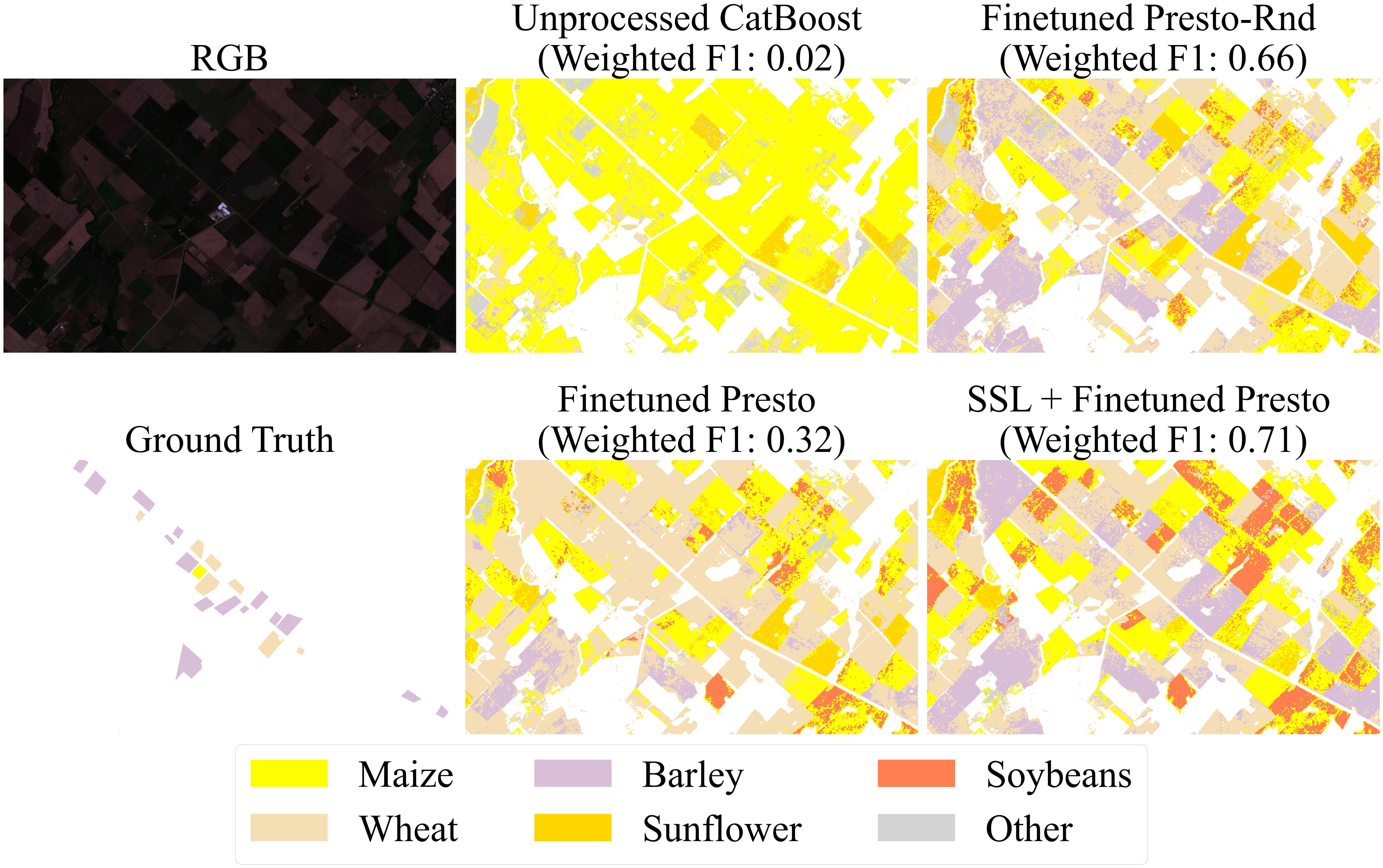}
    \caption{A qualitative assessment of the crop type results. All models were trained using the ``Random'' split before generating these patches. The results of different models are visually very different, with SSL + Finetuned Presto model providing better overall F1 score (computed on a limited available ground truth for the patch). Patch location: Argentina.}
    \label{fig:croptype_argentina_patch}
\end{figure}

\begin{figure}
    \centering
    \includegraphics[width=1\linewidth]{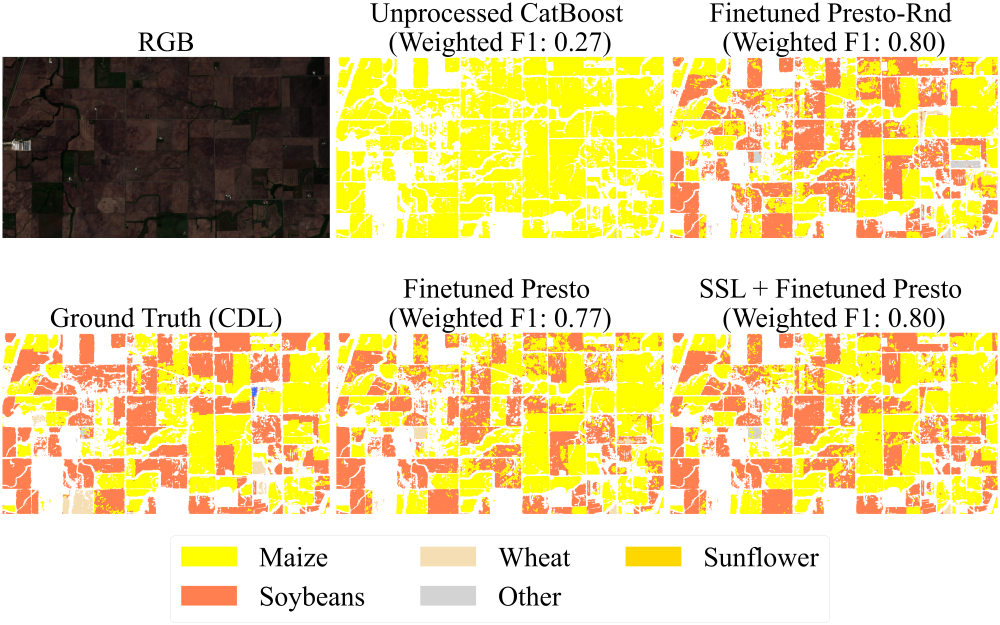}
    \caption{A qualitative assessment of the crop type results. All models were trained using the ``Random'' split before generating these patches. While Unprocessed CatBoost baseline fails completely in this location, all Presto-based models show comparable performance. Patch location: USA.}
    \label{fig:croptype_usa_patch}
\end{figure}

\end{document}